%% file: main.tex
\documentclass[conference]{IEEEtran}
\usepackage{times}

\usepackage[numbers]{natbib}
\usepackage{multicol}
\usepackage[bookmarks=true]{hyperref}
\usepackage{algorithm}
\usepackage{algpseudocode}
\usepackage{graphicx}
\usepackage{amsmath}
\usepackage{amssymb,bm}
\usepackage{booktabs}
\usepackage{tabularx}
\usepackage{xspace}
\usepackage{soul}
\usepackage{cuted}
\usepackage{capt-of}
\usepackage[dvipsnames]{xcolor}
\usepackage{lipsum}

\let\labelindent\relax
\usepackage{enumitem}

\DeclareMathOperator*{\argmax}{arg\,max}
\DeclareMathOperator*{\argmin}{arg\,min}

\makeatletter
\def\blfootnote{\gdef\@thefnmark{}\@footnotetext}
\makeatother

\begin{document}

\title{Dynamic-Resolution Model Learning for\\Object Pile Manipulation}

\author{\authorblockN{Yixuan Wang$^{2*}$\quad Yunzhu Li$^{1,2*}$\quad Katherine Driggs-Campbell$^2$\quad Li Fei-Fei$^1$\quad Jiajun Wu$^1$}
\authorblockA{$^1$Stanford University\quad $^2$University of Illinois Urbana-Champaign\\
{\tt\small \{yixuan22, yunzhuli, krdc\}@illinois.edu, \{feifeili, jiajunwu\}@cs.stanford.edu}}
}

\maketitle
\IEEEpeerreviewmaketitle

\blfootnote{$^*$Denotes equal contribution.}
\blfootnote{$^\dagger$\url{https://RoboPIL.github.io/dyn-res-pile-manip/}}
\input{text/abstract.tex}
\input{text/intro.tex}

\input{text/related.tex}

\input{text/method.tex}

\input{text/experiments.tex}

\input{text/conclusion.tex}

\textit{Acknowledgments:}
This work is in part supported by Stanford Institute for Human-Centered Artificial Intelligence (HAI), Toyota Research Institute (TRI), NSF RI \#2211258, ONR MURI N00014-22-1-2740, and Amazon.

\bibliographystyle{plainnat}
\bibliography{references}

\end{document}


\title{Adaptive Scene Representation Learning in\\World Models for Object Pile Manipulation}

\author{Author Names Omitted for Anonymous Review. Paper-ID 24}

\input{text/supp.tex}

%% file: text/abstract.tex
\input{figText/teaser.tex}

\begin{abstract}

Dynamics models learned from visual observations have shown to be effective in various robotic manipulation tasks. One of the key questions for learning such dynamics models is what scene representation to use. Prior works typically assume representation at a fixed dimension or resolution, which may be inefficient for simple tasks and ineffective for more complicated tasks. In this work, we investigate how to learn dynamic and adaptive representations at different levels of abstraction to achieve the optimal trade-off between efficiency and effectiveness. Specifically, we construct dynamic-resolution particle representations of the environment and learn a unified dynamics model using graph neural networks (GNNs) that allows continuous selection of the abstraction level. During test time, the agent can adaptively determine the optimal resolution at each model-predictive control (MPC) step. We evaluate our method in object pile manipulation, a task we commonly encounter in cooking, agriculture, manufacturing, and pharmaceutical applications. Through comprehensive evaluations both in the simulation and the real world, we show that our method achieves significantly better performance than state-of-the-art fixed-resolution baselines at the gathering, sorting, and redistribution of granular object piles made with various instances like coffee beans, almonds, corn, etc.

\end{abstract}

%% file: figText/teaser.tex
\begin{strip}
    \includegraphics[width=\linewidth]{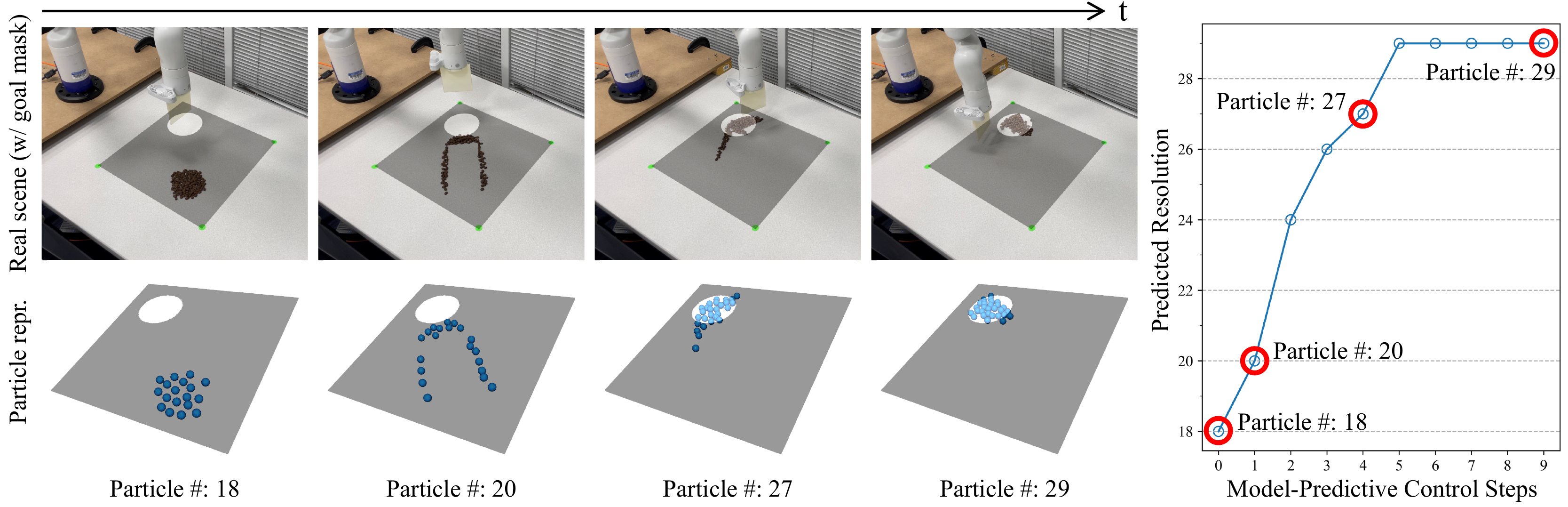}
    \vspace{-18pt}
    \captionof{figure}{\small
    \textbf{Dynamic-Resolution Model Learning for Object Pile Manipulation in the Real World.}
    Depending on the progression of a task, representations at different granularity levels may be needed at each model-predictive control (MPC) step to make the most effective progress on the overall task.
    In this work, we construct dynamic-resolution particle representations of the environment and learn a \textit{unified} dynamics model using graph neural networks~(GNNs) that allows adaptive selection of the abstraction level.
    In this figure, we demonstrate a real-world task of gathering the object pile into a target region. Figures on the left show the task execution process and the corresponding particle representation. The plot on the right shows the predicted optimal resolution at each MPC step, where the \textcolor{red}{red circles} correspond to the frames on the left.
    For video illustrations, we invite you to visit our project page$^\dagger$.
    }
    \vspace{-5pt}
    \label{fig:teaser}
\end{strip}

%% file: text/intro.tex
\section{Introduction}

Predictive models have been one of the core components in various robotic systems, including navigation~\cite{ivanovic2020mats}, locomotion~\cite{kuindersma2016optimization}, and manipulation~\cite{hogan2016feedback,zhou2019pushing}. For robotic manipulation in particular, people have been learning dynamics models of the environment from visual observations and demonstrated impressive results in various manipulation tasks~\cite{finn2017deep,lin2022learning,wu2022daydreamer,shi2022robocraft}.
A learning-based dynamics model typically involves an encoder that maps the visual observation to a scene representation, and a predictive model predicts the representation's evolution given an external action. Different choices of scene representations (e.g., latent vectors~\cite{hafner2019learning,hafner2019dream,li20223d}, object-centric~\cite{yi2019clevrer,driess2022learning} or keypoint representations~\cite{minderer2019unsupervised,manuelli2020keypoints,wang2022dynamical}) imply different expressiveness and generalization capabilities. Therefore, it is of critical importance to think carefully about what scene representation to use for a given task.

Prior works typically use a fixed representation throughout the entire task. However, to achieve the best trade-off between efficiency and effectiveness, the optimal representation may need to be different depending on the object, the task, or even different stages of a task. An ideal representation should be minimum in its capacity (i.e., \textit{efficiency}) but sufficient to accomplish the downstream tasks (i.e., \textit{effectiveness})~\cite{tishby2000information,bengio2013representation}.
Take the object pile manipulation task as an example. When the task objectives are different, the more complicated target configuration will require a more fine-grained model to capture all the details. On the other hand, if the targets are the same, depending on the progression of the task, we might want representations at different abstraction levels to come up with the most effective action, as illustrated in Figure~\ref{fig:teaser}.

In this work, we focus on the robotic manipulation of object piles, a ubiquitous task critical for deploying robotic manipulators in cooking, agriculture, manufacturing, and pharmaceutical scenarios.
Object pile manipulation is challenging in that the environment has extremely high degrees of freedom~\cite{schenck2017learning}. Therefore, developing methods that solve this complex task allows us to best demonstrate how we can learn dynamics models at different levels of abstraction to achieve the optimal trade-off between efficiency and effectiveness. 

Our goal is to learn a \textit{unified} dynamics model that can adaptively express the world at different granularity levels, from which the agent can automatically determine the optimal resolution given the task objective and the current observation.
Specifically, we introduce a resolution regressor that predicts the optimal resolution conditioned on the current observation and the target configuration. The regressor is learned in a self-supervised fashion with labels coming from Bayesian optimization~\cite{garnett2023bayesian} that determines the most effective resolution for minimizing the task objective under a given time budget.
Besides the resolution regressor, our model also includes perception, dynamics, and planning modules (Figure~\ref{fig:model}). 

During task execution, we follow a model-predictive control (MPC) framework. At each MPC step, the resolution regressor predicts the resolution most effective for control optimization. The perception module then samples particles from the RGBD visual observation based on the predicted resolution. The derived particle-based scene representation, together with the robot action, will be the input to the dynamics model to predict the environment's evolution. The dynamics model can then be used for trajectory optimization to derive the action sequence.
Specifically, the dynamics model is instantiated as a graph neural network consisting of node and edge encoders. Such compositional structures naturally generalize to particle sets of different sizes and densities---a unified graph-based dynamics model can support model-predictive control at various abstraction levels, selected continuously by the resolution regressor.

We evaluate the model in various object pile manipulation tasks, including gathering spread-out pieces to a specific location, redistributing the pieces into complicated target shapes, and sorting multiple object piles. The tasks involve the manipulation of piles consisting of different instances, including corn kernels, coffee beans, almonds, and candy pieces (Figure~\ref{fig:setup}b).
We show that our model can automatically determine the resolution of the scene representation conditioned on the current observation and the task goal, and make plans to accomplish these tasks.

We make three core contributions: (1) We introduce a framework that, at each planning step, can make continuous predictions to dynamically determine the scene representation at different abstraction levels.
(2) We conduct comprehensive evaluations and suggested that our dynamic scene representation selection performs much better than the fixed-resolution baselines.
(3) We develop a unified robotic manipulation system capable of various object pile manipulation tasks, including gathering, sorting, and redistributing into complicated target configurations.

%% file: text/related.tex
\section{Related Work}

\subsection{Scene Representation at Different Abstraction Levels}

To build multi-scale models of the dynamical systems, prior works have adopted wavelet-based methods and windowed Fourier Transforms to perform multi-resolution analysis~\cite{daubechies1992ten,debnath2002wavelet,kutz2013data}.
\citet{kevrekidis2003equation,kevrekidis2004equation} investigated equation-free, multi-scale modeling methods via computer-aided analysis.
\citet{kutz2016multiresolution} also combined multi-resolution analysis with dynamic mode decomposition for the decomposition of multi-scale dynamical data. Our method is different in that we directly learn from vision data for the modeling and planning of real-world manipulation systems.

In computer vision, \citet{marr2010vision} laid the foundation by proposing a multi-level representational framework back in 1982.
Since then, people have investigated pyramid methods in image processing~\cite{adelson1984pyramid,burt1987laplacian} using Gaussian, Laplacian, and Steerable filters. Combined with deep neural networks, the multi-resolution visual representation also showed stunning performance in various visual recognition tasks~\cite{he2015spatial,zhao2017pyramid}.
In the field of robotics, reinforcement learning researchers have also studied task- or behavior-level abstractions and come up with various hierarchical reinforcement learning algorithms~\cite{nicolescu2002hierarchical,barto2003recent,botvinick2012hierarchical,vezhnevets2017feudal,nachum2018data,fazeli2019see,pateria2021hierarchical}.
Our method instead focuses on spatial abstractions from vision, where we learned structured representations based on particles to model the object interactions within the environment at different levels.

\subsection{Compositional Model Learning for Robotic Manipulation}

Physics-based models have demonstrated their effectiveness in many robotic manipulation tasks (e.g.,~\cite{hogan2016feedback,zhou2019pushing,pang2022global,suh2022bundled}). However, they typically rely on complete information about the environment, limiting their use in scenarios where full-state estimation is hard or impossible to acquire (e.g., precise shape and pose estimation of each one of the object pieces in Figure~\ref{fig:teaser}).
Learning-based approaches provide a way of building dynamics models directly from visual observations. Prior methods have investigated various scene representations for dynamics modeling and manipulation of objects with complicated physical properties, including clothes~\cite{lin2022learning,huang2022mesh}, ropes~\cite{chang2020model,mitrano2021learning}, fluids~\cite{li20223d}, softbodies~\cite{shen2022acid}, and plasticine~\cite{shi2022robocraft}.
Among the methods, graph-structured neural networks (GNNs) have shown great promise by introducing explicit relational inductive biases~\cite{battaglia2018relational}. Prior works have shown GNNs' effectiveness in modeling compositional dynamical systems involving the interaction between multiple objects~\cite{battaglia2016interaction,chang2016compositional,li2019propagation,sanchez2018graph,funk2022learn2assemble,silver2021planning}, systems represented using particles or meshes~\cite{mrowca2018flexible,li2018learning,ummenhofer2020lagrangian,sanchez2020learning,pfaff2020learning}, or for compositional video prediction~\cite{ye2019compositional,hsieh2018learning,watters2017visual,yi2019clevrer,qi2020learning,tung20203d,zhu2018object}.
However, these works typically assume scene representation at a fixed resolution, whereas our method learns a unified graph dynamics model that can generalize to scene representations at different levels of abstraction.

\input{figText/model.tex}

\subsection{Object Pile Manipulation}

Robotic manipulation of object piles and granular pieces has been one of the core capabilities if we want to deploy robot systems for complicated tasks like cooking and manufacturing. \citet{suh2020surprising} proposed to learn visual dynamics based on linear models for redistributing the object pieces. Along the lines of learning the dynamics of granular pieces, \citet{tuomainen2022manipulation} and \citet{schenck2017learning} also proposed the use of GNNs or convolutional neural dynamics models for scooping and dumpling granular pieces.
Other works introduced success predictors for excavation tasks~\cite{lu2021excavation}, a self-supervised mass predictor for grasping granular foods~\cite{takahashi2021uncertainty}, visual serving for shaping deformable plastic materials~\cite{cherubini2020model}, or data-driven methods to calibrate the physics-based simulators for both manipulation and locomotion tasks~\cite{zhu2019data,matl2020inferring}. Audio feedback has also shown to be effective at estimating the amount and flow of granular materials~\cite{clarke2018learning}.
Our work instead focuses on three tasks (i.e., gather, redistribute, and sort object pieces) using a unified dynamic-resolution graph dynamics to balance efficiency and effectiveness for real-world deployment.

%% file: figText/model.tex
\begin{figure*}[t]
    \centering
    \includegraphics[width=\linewidth]{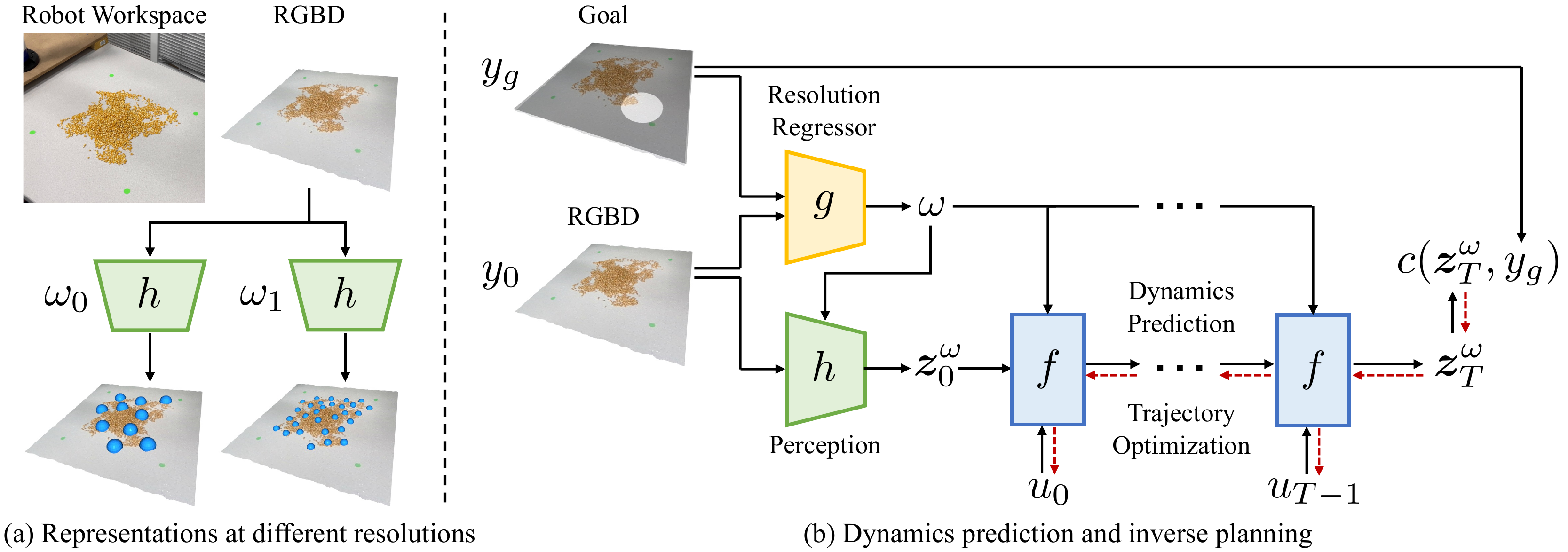}
    \vspace{-18pt}
    \caption{\small
    \textbf{Overview of the proposed framework.}
    (a) Our perception module $h$ processes the input RGBD image and generates particle representations at different levels of abstraction depending on the resolution $\omega$.
    (b) The resolution regressor $g$ takes the current observation $y_0$ and the goal $y_g$ as input. It then predicts the resolution $\omega$ we intend to represent the environment. The dynamics model $f$, conditioned on the dynamically-selected resolution $\omega$ and the input action $u_t$, predicts the temporal evolution of the scene representation $\bm{z}^\omega_t$. During planning time, we calculate the task objective $c(\bm{z}^\omega_T, y_g)$ and backpropagate the gradients to optimize the action sequence $\{ u_t \}$.
    }
    \vspace{-8pt}
    \label{fig:model}
\end{figure*}

%% file: text/method.tex
\section{Method}

In this section, we first present the overall problem formulation. We then discuss the structure of our dynamic-resolution dynamics models, how we learn a resolution regressor to automatically select the scene representation, and how we use the model in a closed loop for the downstream planning tasks.

\subsection{Problem Formulation}

Our goal is to derive the resolution $\omega$ to represent the environment to achieve the best trade-off between efficiency and effectiveness for control optimization. We define the following trajectory optimization problem over a horizon~$T$:
\begin{equation}
\begin{aligned}
\min_{\{u_t\}} \quad &c(\bm{z}^\omega_T, y_g), \\
\textrm{s.t.} \quad &\omega = g(y_0, y_g), \\
&\bm{z}^\omega_0 = h(y_0, \omega), \\
&\bm{z}^\omega_{t+1} = f(\bm{z}^\omega_t, u_t, \omega), \\
\label{eq:obj_fn}
\end{aligned}
\vspace{-15pt}
\end{equation}
where the resolution regressor $g(\cdot,\cdot)$ takes the current observation $y_0$ and the goal configuration $y_g$ as input and predicts the model resolution. $h(\cdot, \cdot)$, the perception module, takes in the current observation $y_0$ and the predicted resolution $\omega$, then derives the scene representation $\bm{z}^\omega_0$ for the current time step. The dynamics module $f(\cdot, \cdot, \cdot)$ takes the current scene representation $\bm{z}^\omega_t$, the input action $u_t$, and the resolution $\omega$ as inputs, and then predicts the representation's evolution at the next time step $\bm{z}^\omega_{t+1}$.
The optimization aims to find the action sequence $\{ u_t \}$ to minimize the task objective $c(\bm{z}_T^\omega, y_g)$.

In the following sections, we describe (1) the details of the perception module $h(\cdot, \cdot)$ and the dynamics module $f(\cdot, \cdot, \cdot)$ in Section~\ref{sec:gnn}, (2) how we obtain the self-supervision for the resolution regressor $g(\cdot,\cdot)$ in Section~\ref{sec:ada_res}, and (3) how we solve Equation~\ref{eq:obj_fn} in a closed planning loop in Section~\ref{sec:planning}.

\subsection{Dynamic-Resolution Model Learning Using GNNs}
\label{sec:gnn}

To instantiate the optimization problem defined in Equation~\ref{eq:obj_fn}, we use graphs of different sizes as the representation $\bm{z}^\omega_t = (\mathcal{O}_t, \mathcal{E}_t)$, where $\omega$ indicates the number of vertices in the graph. The vertices $\mathcal{O}_t=\{o^i_t\}_{i=1,\dots,|\mathcal{O}_t|}$ denote the particle set and $o^i_t$ represents the 3D position of the $i^\text{th}$ particle. The edge set $\mathcal{E}_t = \{e^j_t\}_{j=1,\dots,|\mathcal{E}_t|}$ denotes the relations between the particles, where $e^j_t = (u^j_t, v^j_t)$ denotes an edge pointing from particle of index $v^j_t$ to $u^j_t$.

To obtain the particle set $\mathcal{O}_t$ from the RGBD visual observation $y_t$, we first transform the RGBD image into a point cloud and then segment the point cloud to obtain the foreground according to color and depth information $\bar{y}_t \in \mathbb{R}^{N \times 3}$. We then deploy the farthest point sampling technique~\cite{moenning2003fast} to subsample the foreground but ensure sufficient coverage of $\bar{y}_t$. Specifically, given already sampled particles $o_t^{1,\dots,i-1}$, we apply
\begin{equation}
\begin{aligned}
    o^i_t = \argmax_{y^k\in \bar{y}_t}\min_{o^j_t\in o_t^{1, \dots, i-1}}\|y^k - o^j_t\|^2_2
\end{aligned}
\end{equation}
to find the $i^\text{th}$ particle $o^i_t$. We iteratively apply this process until we reach $\omega$ particles. Different choices of $\omega$ indicate scene representations at different abstraction levels, as illustrated in Figure~\ref{fig:model}a.
The edge set is constructed dynamically over time and connects particles within a predefined distance while limiting the maximum number of edges a node can have.

We instantiate the dynamics model $f(\cdot, \cdot, \cdot)$ as graph neural networks (GNNs) that predict the evolution of the graph representation $\bm{z}^\omega_t$ under external actions $u_t$ and the selected resolution $\omega$. $f(\cdot, \cdot, \cdot)$ consists of node and edge encoders $f^\text{enc}_{\mathcal{O}}(\cdot,\cdot,\cdot)$, $f^\text{enc}_{\mathcal{E}}(\cdot,\cdot,\cdot)$ to obtain node and edge representations:
\begin{equation}
\begin{aligned}
& p^i_t = f^\text{enc}_{\mathcal{O}}(o_t^i, u_t, \omega), \quad i = 1,\dots, |\mathcal{O}_t|, \\
& q^j_t = f^\text{enc}_{\mathcal{E}}(o_t^{u^j_t}, o_t^{v^j_t},\omega), \quad j = 1, \dots, |\mathcal{E}_t|.
\end{aligned}
\end{equation}
We then have node and edge decoders $f^\text{dec}_{\mathcal{O}}(\cdot,\cdot)$, $f^\text{dec}_{\mathcal{E}}(\cdot,\cdot)$ to obtain the corresponding representations and predict the representation at the next time step:
\begin{equation}
\begin{aligned}
& r^j_t = f^\text{dec}_\mathcal{E}(q^j_t,\omega), \quad j=1,\dots,|\mathcal{E}_t|,\\
& \hat{o}^i_{t+1} = f^\text{dec}_\mathcal{O}(p^i_t, \sum_{j\in\mathcal{N}_i}r^j_t), \quad i=1,\dots, |\mathcal{O}_t|,
\end{aligned}
\end{equation}
where $\mathcal{N}_i$ is the index set of the edges, in which particle $i$ is the receiver. In practice, we follow \citet{li2019propagation} and use multi-step message passing over the graph to approximate the instantaneous propagation of forces.

To train the dynamics model, we iteratively predict future particle states over a time horizon of $T$ and then optimize the neural network's parameters by minimizing the mean squared error (MSE) between the predictions and the ground truth future states:
\begin{equation}
\begin{aligned}
\mathcal{L} = \frac{1}{T\cdot|\mathcal{O}_t|} \sum_{t'=1}^T \sum_{i=1}^{|\mathcal{O}_t|}\| \hat{o}^i_{t+t'} - o^i_{t+t'} \|^2_2.
\label{eq:obj_dy}
\end{aligned}
\end{equation}

\subsection{Adaptive Resolution Selection via Self-Supervised Learning}
\label{sec:ada_res}

The previous sections discussed how to obtain the particle set and how we predict its evolution given a resolution $\omega$. In this section, we present how we learn the resolution regressor $g(\cdot, \cdot)$ in Equation~\ref{eq:obj_fn} that can automatically determine the resolution in a self-supervised manner.
Specifically, we intend to find the resolution $\omega$ that is the most effective for minimizing the task objective given the current observation $y_0$ and the goal $y_g$. We reformulate the optimization problem in Equation~\ref{eq:obj_fn} by considering $\omega$ as a variable of the objective function as the following:
\begin{equation}
\begin{aligned}
c^*(y_0, y_g, \omega) = \min_{\{u_t\}} \quad &c(\bm{z}^\omega_T, y_g), \\
\textrm{s.t.} \quad &\bm{z}^\omega_0 = h(y_0, \omega), \\
& \bm{z}^\omega_{t+1} = f(\bm{z}^\omega_t, u_t, \omega).
\label{eq:obj_fn_fix_omega}
\end{aligned}
\end{equation}
For a given $\omega$, we solve the above optimization problem via a combination of sampling and gradient descent using shooting methods~\cite{tedrake2009underactuated} under a given time budget---the higher resolution representation will go through fewer optimization iterations. For simplicity, we denote the objective in Equation~\ref{eq:obj_fn_fix_omega} as $c^*(\omega)$ in the following part of this section.

Given the formulation, we are then interested in finding the parameter $\omega$ that can minimize the following objective:
\begin{equation}
\begin{aligned}
\min_{\omega}\quad & c^+(\omega) = c^*(\omega) + R(\omega), \\
\textrm{s.t.}\quad & \omega \in (\omega_{\textrm{min}}, \omega_{\textrm{max}}),
\label{eq:opt_omega}
\end{aligned}
\end{equation}
where $R(\omega)$ is a regularizer penalizing the choice of an excessively large $\omega$ to encourage efficiency. Regularizer details can be found in supplementary materials.
We use Bayesian optimization~\cite{snoek2012practical} to find the optimal $\omega$ by iteratively sampling $\omega$ and approximating $c^+(\omega)$ using the Gaussian process.
At each sampling stage, we sample one or more data points $\omega_i$ according to the expected improvement of the objective function and evaluate their value $c^+(\omega_i)$.
Then, at the approximation stage, we assume the distribution of $c^+(\omega)$ follows the Gaussian distribution $\mathcal{N}(\mu(\omega), \sigma^2)$; thus, the joint distribution of the evaluated points $\mathbf{\Omega}_\text{train} = [\omega_1, \dots, \omega_n]$ and the testing points $\mathbf{\Omega}_\text{test} = [\omega_1', \dots, \omega_m']$ can be expressed as the following:
\begin{equation}
\begin{aligned}
\mathbf{C}_\text{train} &= [c^+(\omega_1), \dots, c^+(\omega_n)], \\
\mathbf{M}_\text{train} &= [\mu(\omega_1), \dots, \mu(\omega_n)], \\
\mathbf{C}_\text{test} &= [c^+(\omega_1'), \dots, c^+(\omega_m')], \\
\mathbf{M}_\text{test} &= [\mu(\omega_1'), \dots, \mu(\omega_m')], \\
\begin{bmatrix}
    \mathbf{C}_\text{train}\\
    \mathbf{C}_\text{test}
\end{bmatrix}
&\sim
\mathcal{N}\left(
\begin{bmatrix}
    \mathbf{M}_\text{train} \\
    \mathbf{M}_\text{test}
\end{bmatrix},
\begin{bmatrix}
    \mathbf{K} & \mathbf{K_*} \\
    \mathbf{K}_*^\top & \mathbf{K_{**}}
\end{bmatrix}
\right),
\label{eq:joint_distri}
\end{aligned}
\end{equation}
where $\mathbf{K}$ is a kernel function matrix derived via $\mathbf{K}=K(\mathbf{\Omega}_\text{train}, \mathbf{\Omega}_\text{train})$. $K(\cdot,\cdot)$ is the kernel function used to compute the covariance. Similarly, $\mathbf{K_*}=K(\mathbf{\Omega}_\text{train}, \mathbf{\Omega}_\text{test})$ and $\mathbf{K_{**}}=K(\mathbf{\Omega}_\text{test}, \mathbf{\Omega}_\text{test})$.

Equation \ref{eq:joint_distri} shows the joint probability of $\mathbf{C}_\text{train}$ and  $\mathbf{C}_\text{test}$ conditioned on $\mathbf{\Omega}_\text{train}$ and $\mathbf{\Omega}_\text{test}$.
Through marginalization, we could fit $c^+(\omega)$ using the following conditional distribution:
\begin{equation}
\begin{aligned}
\mathbf{C}_\text{test} &| \mathbf{C}_\text{train}, \mathbf{\Omega}_\text{train}, \mathbf{\Omega}_\text{test} \sim \\
&\mathcal{N}(\mathbf{K}_*^\top \mathbf{KC}_\text{train}, \mathbf{K}_{**} - \mathbf{K}_*^\top\mathbf{K}^{-1}\mathbf{K_*} ).
\label{eq:gp}
\end{aligned}
\end{equation}
We can then use the mean value of the Gaussian distribution in Equation~\ref{eq:gp} as the metric to minimize $c^+(\omega)$. Therefore, the solution to Equation~\ref{eq:opt_omega} is approximated as the following:
\begin{equation}
\begin{aligned}
\omega^* = \argmin_{\omega} \quad & \mathbf{K}_*^\top \mathbf{KC}_\text{train} \\
\textrm{s.t.}\quad & \omega \in (\omega_{\text{min}}, \omega_{\text{max}}).
\label{eq:min_omega}
\end{aligned}
\end{equation}

To train the resolution regressor, we randomly generate a dataset containing the observation and goal pairs $(y_0, y_g)$. For each pair, we follow the above optimization process to generate the optimal resolution label $\omega^*$. We then train the resolution regressor $\omega = g(y_0,y_g)$ to predict the resolution based on the observation and the goal via supervised learning.
Training the $\omega$ regressor is a self-supervised learning process, as the labels are automatically generated via an optimization process without any human labeling.

\subsection{Closed-Loop Planning via Adaptive Repr. Selection}
\label{sec:planning}

Now that we have obtained the resolution regressor $g$, the perception module $h$, and the dynamics module $f$. We can wire things together to solve Equation~\ref{eq:obj_fn} and use the optimized action sequence in a closed loop within a model-predictive control (MPC) framework~\cite{camacho2013model}.
Specifically, for each MPC step, we follow Algorithm~\ref{alg:planning}, which first determines the resolution to represent the environment, then uses a combination of sampling and gradient descent to derive the action sequence through trajectory optimization using the shooting method.
We then execute the first action from the action sequence in the real world, obtain new observations, and apply Algorithm~\ref{alg:planning} again. Such a process allows us to take feedback from the environment and adaptively select the most appropriate resolution at each step as the task progresses.
Figure~\ref{fig:model}b also shows an overview of the future prediction and inverse planning process.
Details including task objective definition and MPC hyperparameter are included in supplementary materials.

\begin{center}
\begin{algorithm}[h!]
   \caption{Trajectory optimization at each MPC step}
   \label{alg:planning}
   \begin{algorithmic}
   \State {\bfseries Input:} Current observation $y_0$, goal $y_g$, time horizon $T$,\\
     \quad the resolution regressor $g$,  the perception module $h$, \\
     \quad the dynamics module $f$, and gradient descent iteration $N$
   \State \textbf{Output: } Actions $u_{0:T-1}$\\

   \State Predict the resolution $\omega \gets g(y_0, y_g)$
   \State Obtain the current representation $\bm{z}^\omega_0 \gets h(y_0, \omega)$
   \State Sample $M$ action sequences $\hat{u}^{1:M}_{0:T-1}$
   \For {$m=1,\dots,M$}
       \For {$i=1,\dots,N$}
           \For {$t=0,\dots,T-1$}
               \State Predict the next step $\bm{z}^\omega_{t+1} \gets f(\bm{z}^\omega_{t}, \hat{u}^m_t, \omega)$
           \EndFor
           \State Calculate the task loss $c^m \gets c(\bm{z}^\omega_T, y_g)$
           \If {$i < N$}
               \State Update $\hat{u}^m_{0:T-1}$ using gradients $\nabla_{\hat{u}^m_{0:T-1}}c^m$
           \EndIf
       \EndFor
   \EndFor

   \State $m^* \gets \argmin_m c^m$
   \State Return $\hat{u}^{m^*}_{0:T-1}$
\end{algorithmic}
\end{algorithm}
\end{center}

%% file: text/experiments.tex
\section{Experiments}

In this section, we evaluate the proposed framework in various object pile manipulation tasks. In particular, we aim to answer the following three questions through the experiments. (1) Does a trade-off exist between efficiency and effectiveness as we navigate through representations at different abstraction levels? (2) Is a fixed-resolution dynamics model sufficient, or do we need to dynamically select the resolution at each MPC step? (3) Can our dynamic-resolution model accomplish three challenging object pile manipulation tasks: \textbf{Gather}, \textbf{Redistribute}, and \textbf{Sort}?

\input{figText/setup.tex}

\input{figText/tradeoff.tex}

\subsection{Setup}

We conduct experiments in both the simulation environment and the real world. The simulation environment is built using NVIDIA FleX~\cite{macklin2014unified,li2018learning}, a position-based simulator capable of simulating the interactions between a large number of object pieces.
In the real world, we conducted experiments using the setup shown in Figure~\ref{fig:setup}a. We use RealSense D455 as the top-down camera to capture the RGBD visual observations of the workspace. We attach a flat pusher to the robotic manipulator's end effector to manipulate the object piles.

\subsection{Tasks}
We evaluate our methods on three object pile manipulation tasks that are common in daily life.
\begin{itemize}
    \item \textbf{Gather}: The robot needs to push the object pile into a target blob with different locations and radii.
    \item \textbf{Redistribute}: The robot is tasked to manipulate the object piles into many complex target shapes, such as letters.
    \item \textbf{Sort}: The robot has to move two different object piles to target locations without mixing each other.
\end{itemize}
We use a unified dynamics model for all three tasks, which involve objects pieces of different granularities, appearances, and physical properties (Figure~\ref{fig:setup}b).

\subsection{Trade-Off Between Efficiency and Effectiveness}
\label{sec:q1}

\input{figText/quanti.tex}
\input{figText/quali.tex}

The trade-off between efficiency and effectiveness can vary depending on the tasks, the current, and the goal configurations.
As we have discussed in Section~\ref{sec:ada_res}, given the resolution $\omega$, we set a fixed time budget to solve the optimization problem defined in Equation~\ref{eq:obj_fn_fix_omega}.
Intuitively, if the resolution is too low, the representation will not contain sufficiently detailed information about the environment to accomplish the task, the optimization of which is efficient but not effective enough to finish the task.
On the contrary, if we choose an excessively high resolution, the representation will carry redundant information not necessary for the task and can be inefficient in optimization.
We thus conduct experiments evaluating whether the trade-off exists (i.e., whether the optimal resolution $\omega$ calculated from Equation~\ref{eq:min_omega} is different for different initial and goal configurations).

We use Bayesian optimization and follow the algorithm described in Section~\ref{sec:ada_res} to find the optimal trade-off on \textbf{Gather} and \textbf{Redistribute} tasks in the simulation. As shown in Figure \ref{fig:tradeoff}a, higher-resolution dynamics models do not necessarily lead to better performance due to their optimization inefficiency. Compared between goal configurations, even if the current observation is the same, a more complicated goal typically requires a higher resolution representation to make the most effective task progression.
More specifically, when the target region is a plain circle, the coarse representation captures the rough shape of the object pile, sufficient for the task objective, allowing more efficient optimization than the higher-resolution counterparts.
However, when the target region has a more complicated shape, low-resolution representation fails to inform downstream MPC of detailed object pile shapes. Therefore, high-resolution representation is necessary for effective trajectory optimization.

The desired representation does not only depend on goal configurations.
Even if the goal configurations are the same, different initial configurations can also lead to different optimal resolutions, as illustrated in Figure~\ref{fig:tradeoff}b.
When the initial configuration is more spread out, the most effective way of decreasing the loss is by pushing the outlying pieces to the goal region. Our farthest sampling strategy, even with just a few particles, could capture outlying pieces and helps the agent to make good progress.
Therefore, when pieces are sufficiently spread out, higher particle resolution does not necessarily contain more useful information for the task but makes the optimization process inefficient.
On the other hand, when the initial configuration concentrates on the goal region, to effectively decrease the task objective, MPC needs more detailed information about the object pile's geometry to pinpoint the mismatching area.
For example, the agent needs to know more precise contours of the goal region and the outlying part of object piles to decide how to improve the planning results further.
Low-resolution representations will be less effective in revealing the difference between the current observation and the goal, thus less helpful in guiding the agent to make action decisions.

\subsection{Is a Single Resolution Dynamics Model Sufficient?}
\label{sec:q2}

Although there is a trade-off between representation resolution and task progression, can we benefit from this trade-off in trajectory optimization? We compared our dynamic-resolution dynamics model with fixed-resolution dynamics models on \textbf{Gather} and \textbf{Redistribute} tasks. 
Figure~\ref{fig:teaser} shows how our model changes its resolution prediction as MPC proceeds in the real world. Trained on the generated dataset of optimal $\omega$, our regressor learned that fixing a resolution throughout the MPC process is not optimal. Instead, our regressor learns to adapt the resolution according to the current observation feedback. In addition, for the example shown in Figure~\ref{fig:teaser}, we can see that the resolution increases as object piles approach the goal. This matches our expectation as explained in Section~\ref{sec:q1}.

We quantitatively evaluate different fixed-resolution baselines and our adaptive representation learning algorithms in simulation. We record the final step distribution distance between object piles and the goal.
Specifically, given a distance threshold $\tau_p$, the number of tasks with a distance lower than $\tau_p$ is $N_p$, and the total number of tasks is $N$. The task score is then defined as $N_p/N$ (i.e., y-axis in Figure~\ref{fig:quanti}).
Our adaptive resolution model almost always achieves the highest task score, regardless of the threshold used.

Figure~\ref{fig:quali}a shows a qualitative comparison between the fixed-resolution baselines and our dynamic-resolution selection method on the \textbf{Gather} task in the real world.
All methods start from a near-identical configuration. We can see from the qualitative results that our method manipulates the object pile to a configuration closest to the goal region, whereas the best-performing fixed-resolution baseline still has some outlying pieces far from the goal region.
In addition, we could see from the quantitative evaluation curve in Figure~\ref{fig:quali}b that our model is always the best throughout the whole MPC process.
Representations with an excessively high resolution are unlikely to converge to a decent solution within the time budget, as demonstrated by resolutions 75 and 100.
Conversely, if the representation is too low resolution, it will converge to a loss much higher than our model. A resolution of 25 reached a comparable final loss to our method. However, because the same resolution was ineffective for initial timesteps, its loss does not reduce as rapidly as our adaptive approach.
Because our model could adapt to different resolutions in different scenes, making it more effective at control optimization.

That is why our model could reach the goal region faster than all other fixed-resolution models and consistently performs better at all timestamps, highlighting the benefits of adaptive resolution selection.

\subsection{Can a Unified Dynamics Model Achieve All Three Tasks?}
\label{sec:q3}

We further demonstrate that our method could work on all three tasks and diverse object piles.
For the \textbf{Gather} task, we test our method on different objects with different initial and goal configurations.
From left to right in Figure~\ref{fig:quali}c, our agent gathers different object piles made with almond, granola, or M\&M\textsuperscript{TM}.
Different appearances and physical properties challenge our method's generalization capability.
For example, while almonds and granola are almost quasi-static during the manipulation, M\&M\textsuperscript{TM} will roll around and have high uncertainties in its dynamics.
In addition, unlike almonds and M\&M\textsuperscript{TM}, granola pieces are non-uniform.
Our method has a good performance for all these objects and configurations.

For the \textbf{Redistribute} task, we redistribute carrots and almonds into target letters `J', `T', and `U' with spread-out initial configurations. The final results match the desired letter shape. Please check our supplementary materials for video illustrations of the manipulation process.

For the \textbf{Sort} task, we use a high-level motion planner to find the intermediate waypoints in the image space. Then we use a similar method as \textbf{Gather} task to push the object pile into the target location. For the three examples shown in Figure~\ref{fig:quali}e, we require object piles to go to their own target locations while not mixing with each other.
Objects with different scales and shapes are present here. For example, coffee beans have smaller granularity and round shapes, while candies are relatively large and square. Here we demonstrate success trials of manipulating the object piles to accomplish the \textbf{Sort} task for different objects and goal configurations. Please check our video for the manipulation process.

%% file: figText/setup.tex
\begin{figure}
    \centering
    \includegraphics[width=\linewidth]{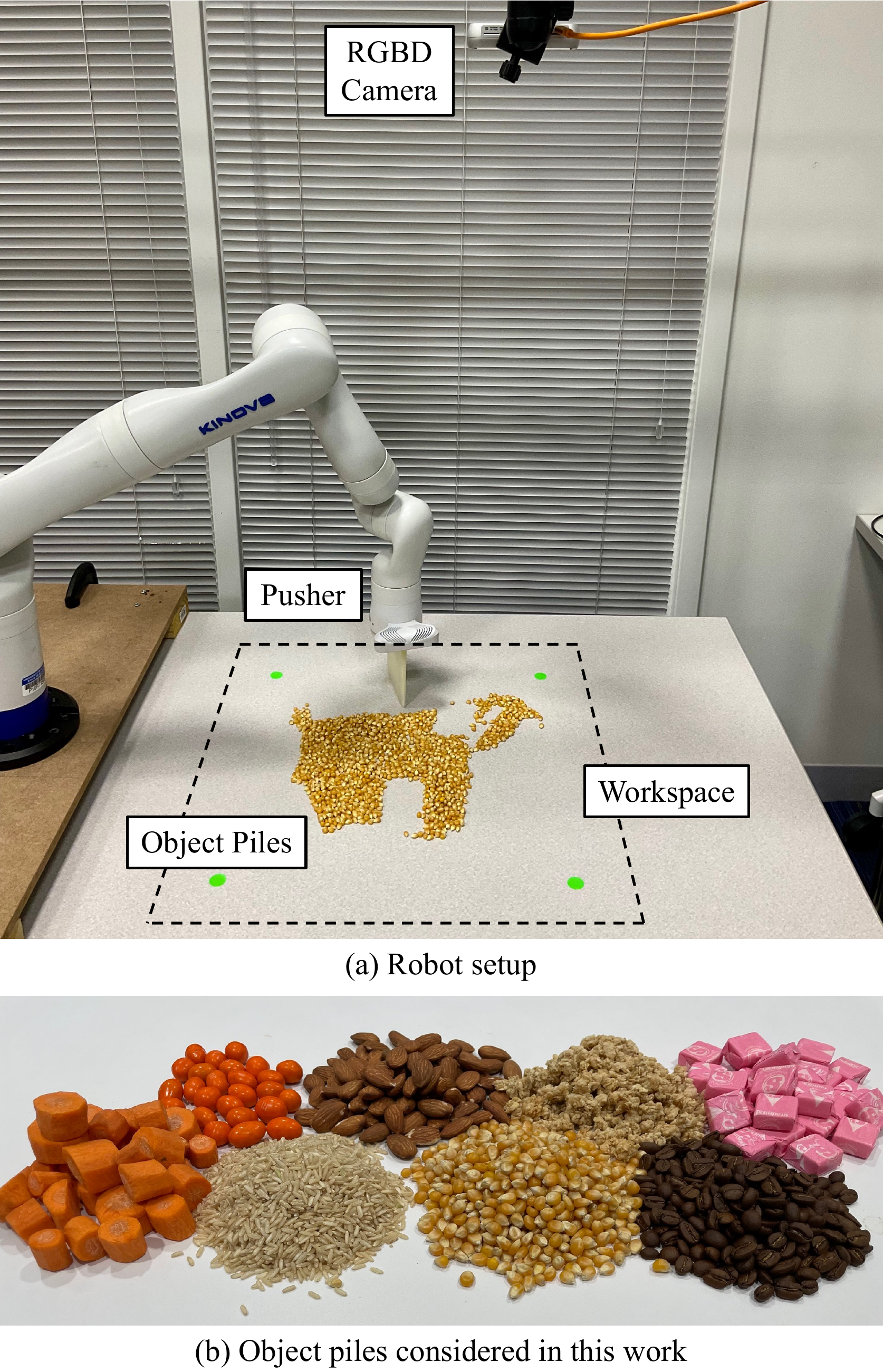}
    \vspace{-18pt}
    \caption{\small
    \textbf{Robot setup and the testing object piles.}
    (a) The dashed black square shows the robot's workspace. The robotic manipulator, equipped with a pusher at the end effector, pushes the object piles within the workspace. A calibrated RGBD camera mounted at the top provides visual observations of the environment. (b) We show the object piles considered in this work, including M\&M, almond, granola, candy, carrot, rice, corn, and coffee beans.
    }
    \vspace{-8pt}
    \label{fig:setup}
\end{figure}

%% file: figText/tradeoff.tex
\begin{figure*}[t]
    \centering
    \includegraphics[width=\linewidth]{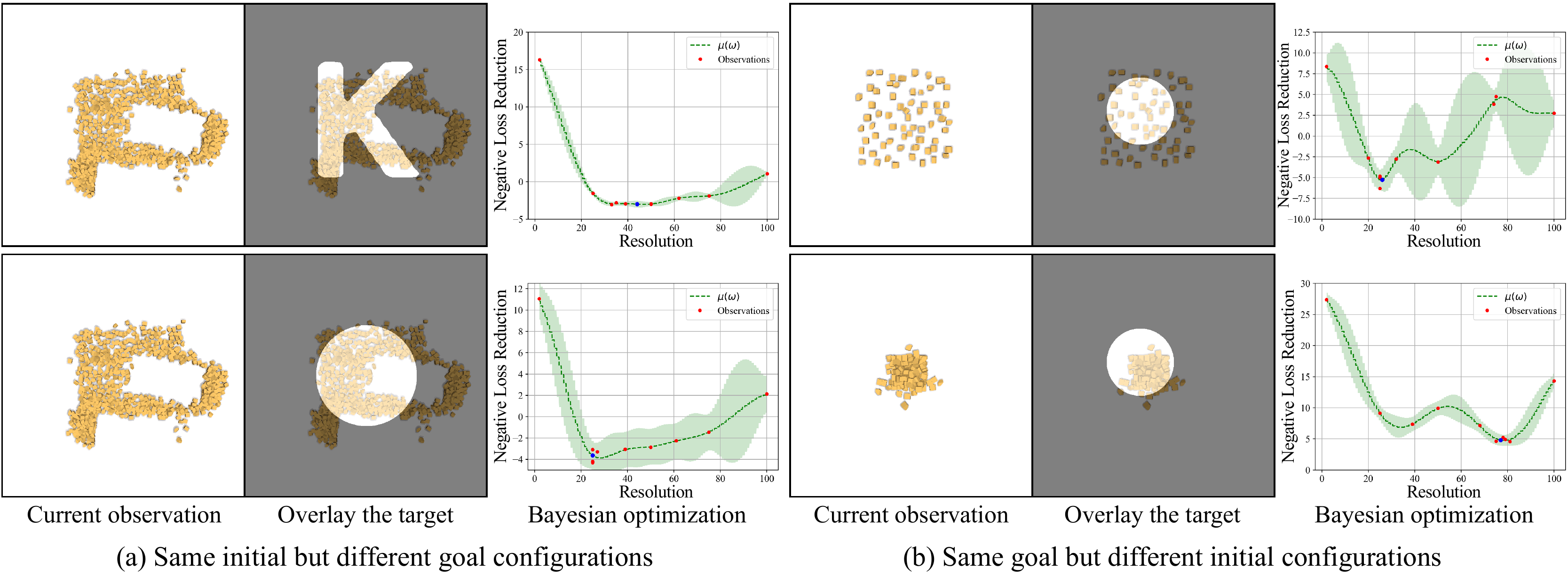}
    \vspace{-18pt}
    \caption{\small
    \textbf{Optimal resolution differs depending on the initial and goal configurations.}
    (a) We show two examples with the same initial but different goal configurations. We apply Bayesian optimization to solve the problem discussed in Section~\ref{sec:ada_res} to find the optimal resolution for both cases. The example with a more complicated target shape requires a higher-resolution representation to be the most effective at making task progress.
    (b) When the goal is to gather the pieces in the center of the workspace, a coarse representation is sufficient for examples with spread-out pieces. The task progresses as long as the agent pushes any outlying pieces toward the goal region. In contrast, a higher-resolution representation is needed to reveal the subtle difference between the initial and goal configurations when they are close.
    }
    \vspace{-8pt}
    \label{fig:tradeoff}
\end{figure*}

%% file: figText/quanti.tex
\begin{figure}[t]
    \centering
    \includegraphics[width=\linewidth]{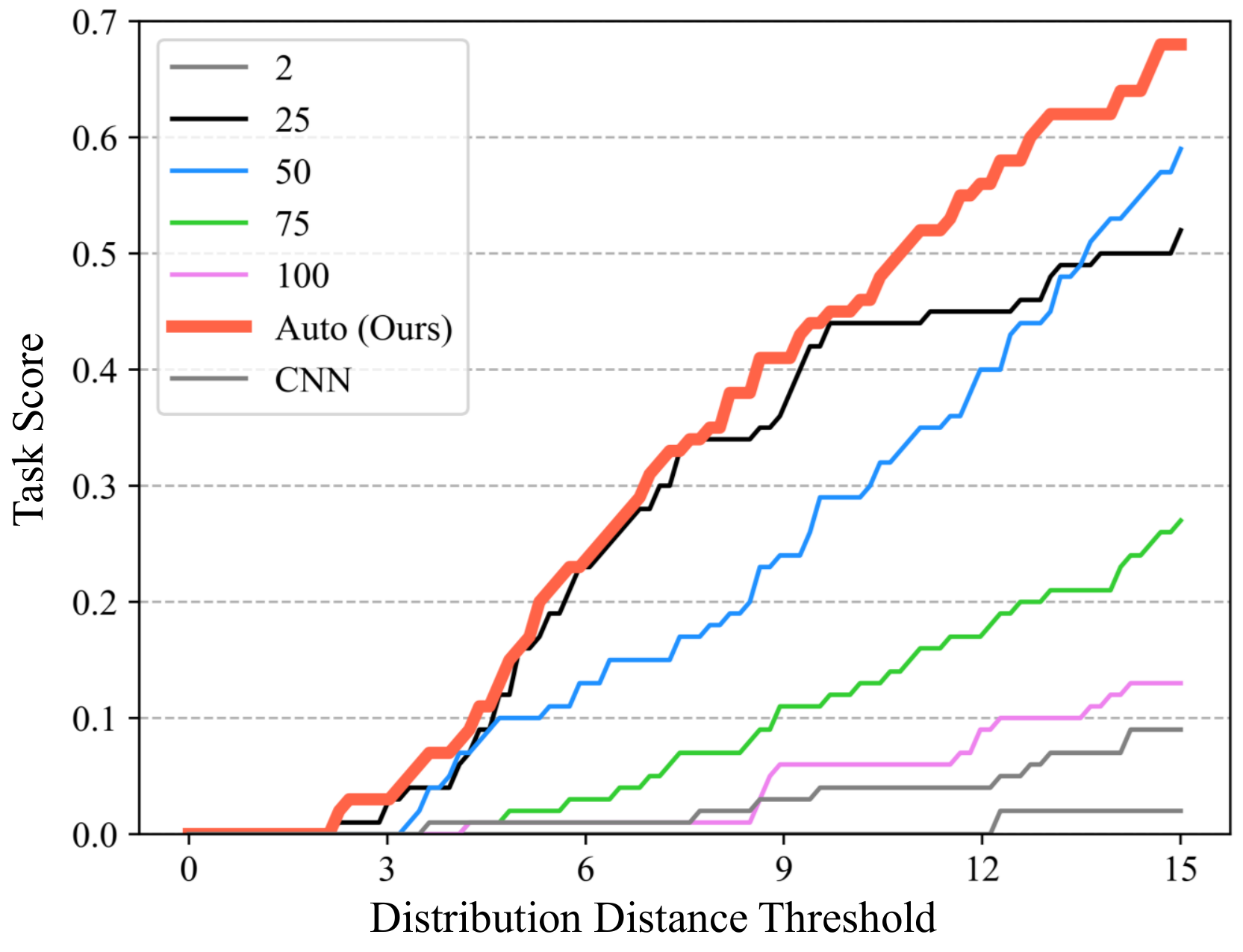}
    \vspace{-18pt}
    \caption{\small
    \textbf{Model-predictive control (MPC) results.}
    We evaluated the MPC performance on different representation choices. We use the task score as the evaluation metric. Task execution trial results in a distribution distance lower than the threshold is considered a success.
    The task score is the number of successful trials divided by the total number of task trials for both the \textbf{Gather} and \textbf{Redistribute} tasks.
    Our method automatically and adaptively selects the scene representation, which achieves the best overall performance compared with the scores of fixed-resolution baselines and a method that uses convolutional neural networks (CNN) as the dynamics model class.
    }
    \vspace{-8pt}
    \label{fig:quanti}
\end{figure}

%% file: figText/quali.tex
\begin{figure*}[h!]
    \centering
    \includegraphics[width=\linewidth]{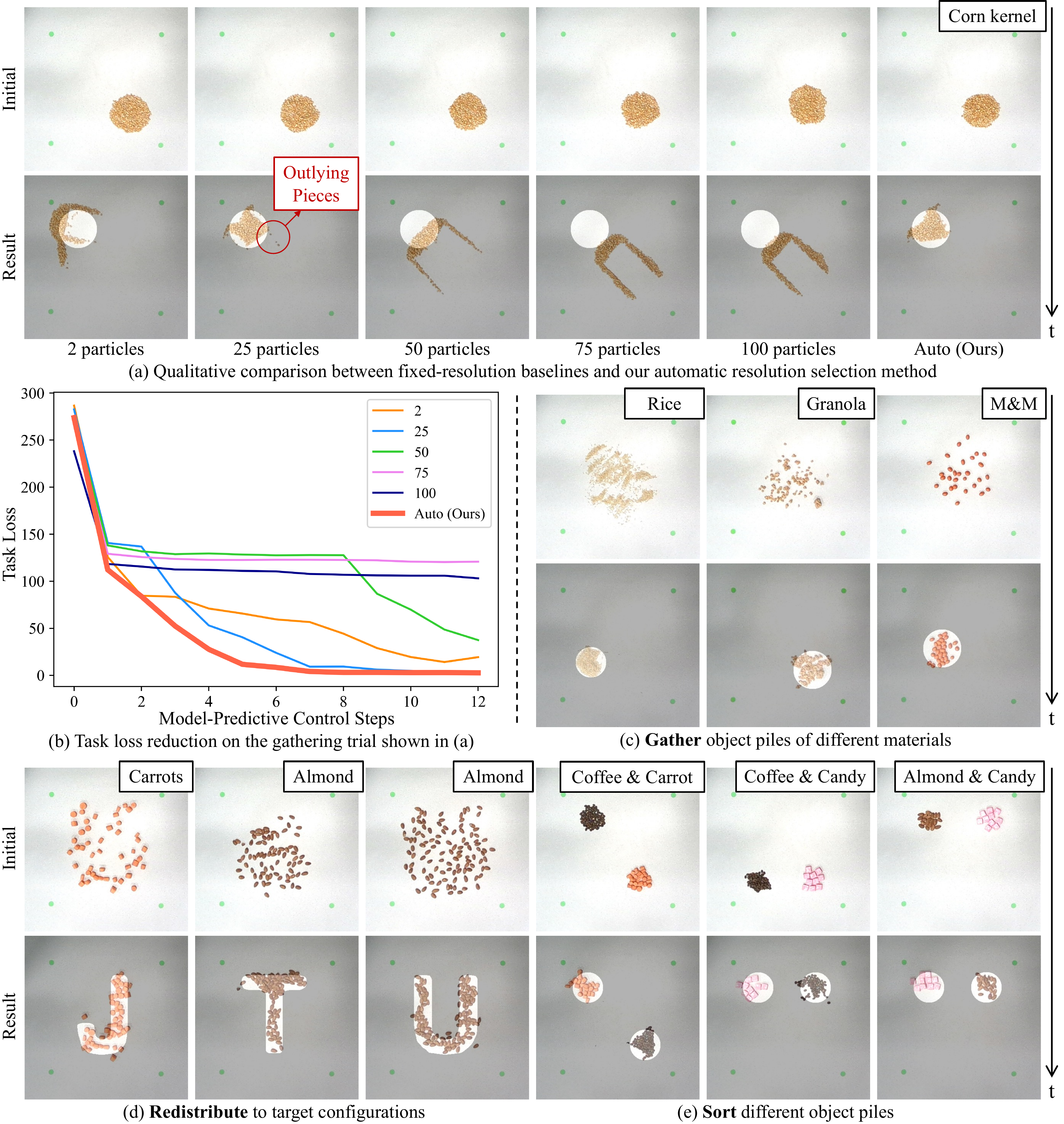}
    \vspace{-18pt}
    \caption{\small
    \textbf{Qualitative results in the real world.}
    (a) Qualitative comparison of MPC performance between our method with fixed-resolution baselines. Our method could gather corn pieces into the target region clearly, whereas the fixed-resolution baselines fail to reach the goal or leave outlying pieces.
    (b) Quantitative comparisons for the qualitative results in (a). Starting from similar initial configurations, our automatic resolution selection method performs the best throughout the MPC steps.
    (c) Evaluation of our method on the \textbf{Gather} task with different objects. The objects vary in their scales and physical properties (e.g., while rice and granola are quasi-static during MPC steps, M\&M can exhibit rolling motions after pushing).
    (d) \textbf{Redistribute} the object pieces into more complicated target configurations.
    Our method can push randomly-spread object piles into the desired letter shapes with clear boundaries.
    (e) Our method can also be coupled with a simple high-level planner to accomplish more complex tasks, such as sorting different object piles into target regions without mixing them.
    }
    \vspace{-8pt}
    \label{fig:quali}
\end{figure*}

%% file: text/conclusion.tex
\section{Conclusion}

Dynamics models play an important role in robotics. Prior works developed dynamics models based on representations of various choices, yet they are typically fixed throughout the entire task.
In this work, we introduced a dynamic and adaptive scene representation learning framework that could automatically find a trade-off between efficiency and effectiveness for different tasks and scenes.
The resolution of the scene representation is predicted online at each time step. And a \textit{unified} dynamics model, instantiated as GNNs, predicts the evolution of the dynamically-selected representation.
The downstream MPC then plans the action sequence to minimize the task objective.
We evaluate our method on three challenging object pile manipulation tasks with diverse initial and goal configurations. We show that our model can dynamically determine the optimal resolution online and has better control performance compared to fixed-resolution baselines.

%% file: text/supp.tex
\section{Method}

\subsection{Perception Module}

Building graph $\bm{z}_t^\omega=(\mathcal{O}_t, \mathcal{E}_t)$ from point cloud $\Bar{y_t} \in \mathbb{R}^{N\times 3}$ contains two phases. First, we sample $\mathcal{O}_t^{\text{fps}}$ from the point cloud using farthest-point sampling. We found that sampled particles from the point cloud are likely to be at the edge of underneath objects. To bias sampled particles towards object centers, we define $\mathcal{O}_t$ as mass centers of $\mathcal{O}_t^{\text{fps}}$ neighbor points. For example, for $o_t^{i, \text{fps}}\in \mathcal{O}_t^{\text{fps}}$, we find the corresponding $o_t^{i}\in \mathcal{O}_t$ by applying
\begin{equation}
\begin{aligned}
\Bar{y_t}' &= \{y|y\in\Bar{y_t}, ||y - o_t^{i, \text{fps}}||_2 \le r_{\text{center}}\} \\
o_t^{i} &= \frac{1}{|\Bar{y_t}'|}\sum_{y\in \Bar{y_t}'}y.
\end{aligned}
\end{equation}
$r_{\text{center}}$ is the hyperparameter to determine the neighbor points.

To find edges $\mathcal{E}_t$, we first approximate particle displacements $\Delta\mathcal{O}_t$ using the action $u_t$. We compute the sweeping region given the action $u_t$. If $o^i_t$ is within the sweeping region, $\Delta o^i_t$ is the vector from $o^i_t$ to the pusher end. We define $\Hat{\mathcal{O}_t} = \Delta\mathcal{O}_t + \mathcal{O}_t$. For $\Hat{o_t}^i$, $(\Hat{o_t}^i, \Hat{o_t}^j) \in \mathcal{E}_t$ if it satisfies the following criteria:
\begin{equation}
\begin{aligned}
||\Hat{o_t}^i - \Hat{o_t}^j||_2 &< r_{\text{edge}} \\
\Hat{o_t}^j &\in \text{kNN}(\Hat{o_t}^i).
\end{aligned}
\end{equation}
$r_{\text{edge}}$ is the hyperparameter to determine the distance needed for two nodes to interact. $\text{kNN}(\Hat{o_t}^i)$ is the set of k-nearest-neighbor of the node $\Hat{o_t}^i$ and $k$ is a hyperparameter.

\subsection{Regressor Module}
We use a regularizer $R(\omega)$ to encourage efficiency as mentioned in Equation 7. $R(\omega)$ is defined as the following:
$$R(\omega)=w_R c(\bm{z}_o, y_g)\omega.$$
$w_R$ is a pre-defined hyperparameter. $c(\bm{z}_0, y_g)$ is the task objective for the current representation. This means that $R(\omega)$ increases linearly as $\omega$ increases. We include $c(\bm{z}_0, y_g)$ in $R(\omega)$ because we observe that $c^*(\omega)$ tends to be larger when the initial task objective is larger.

\subsection{Control Module}
The task objective takes in the representation $\bm{z}_t^\omega=(\mathcal{O}_t, \mathcal{E}_t)$ and binary heatmap $y_g$. Given camera intrinsics, we transform $\mathcal{O}_t$ to $\mathcal{P}_t \in \mathbb{R}^{\omega\times 2}$ in image space. We could also extract points within goal region $\mathcal{Q}_t \in \mathbb{R}^{M\times 2}$ using the binary heatmap $y_g$. Using the farthest-point sampling, we sample $\mathcal{Q}_t' \in \mathbb{R}^{M'\times 2}$ from $\mathcal{Q}_t$. Given these definitions, we compute the task objective using the following equation:
\begin{equation}
\begin{aligned}
\label{eq:task_obj}
&c(\bm{z}_t, y_g) \\
= &\sum_{p_i\in \mathcal{P}_t} \min_{q_j\in \mathcal{Q}_t} ||p_i - q_j||_2 + \sum_{q_j\in \mathcal{Q}_t'} \min_{p_i\in \mathcal{P}_t} ||p_i - q_j||_2.
\end{aligned}
\end{equation}
Instead of using $\mathcal{Q}_t$ in the second term, we use $\mathcal{Q}_t'$ so that the first and second terms can be balanced.

For the MPC hyperparameter in Algorithm 1, we use 20 sampled action sequences with $T=1$. The number of gradient descent iterations is 200.

\section{Experiment}
\subsection{Distribution Distance}
The distribution distance is computed similarly to the task objective. We use the RGBD observation to segment out the foreground object and obtain foreground pixels $\mathcal{F}_t \in \mathbb{R}^{F\times 2}$. Similar to Equation \ref{eq:task_obj}, the distribution distance $d$ is defined using the following equation:
\begin{equation}
\begin{aligned}
\label{eq:task_obj}
d = \sum_{f_i\in \mathcal{F}_t} \min_{q_j\in \mathcal{Q}_t} ||f_i - q_j||_2 + \sum_{q_j\in \mathcal{Q}_t} \min_{f_i\in \mathcal{F}_t} ||f_i - q_j||_2
\end{aligned}
\end{equation}

\subsection{High-Level Planner for \textbf{Sort} Task}
We use the A\textsuperscript{*} search algorithm to find the high-level path. We represent every blob of object piles as a circle with a fixed radius in image space. Therefore, the state for one blob can be represented as its 2D blob center in image space. For the search algorithm, if there are $k$ blobs in the scene, the node is the concatenation of $k$ blob centers. Neighbors of one node are those nodes that can be reached by changing one blob center in a single step with no collision. 

To accelerate motion planning, we divide the image space into a sparse grid. The blob center will only be on the grid. In addition, to encourage a path with fewer steps, we add a constant cost for each path so that a path with fewer steps is preferred.